\documentclass[10pt,journal,compsoc]{IEEEtran}
\usepackage{times}
\usepackage{epsfig}

\usepackage{amssymb}

\usepackage{bm}
\usepackage{booktabs}
\usepackage{float}
\usepackage{cuted}
\usepackage{capt-of}
\usepackage{overpic}
\usepackage{multirow}
\usepackage{lipsum}
\usepackage{rotating}
\usepackage{gensymb}

\ifCLASSOPTIONcompsoc
  \usepackage[nocompress]{cite}
\else
  \usepackage{cite}
\fi
\usepackage{graphicx}

\usepackage{amsmath}
\usepackage{array}
\usepackage{url}

\usepackage[dvipsnames]{xcolor}

\begin{document}

\title{PICA: Physics-Integrated Clothed Avatar}
\author{Bo~Peng,
        Yunfan~Tao,
        Haoyu~Zhan,
        Yudong~Guo,
        and~Juyong~Zhang$^\dagger$
\IEEEcompsocitemizethanks{\IEEEcompsocthanksitem B. Peng, Y. Tao, H. Zhan, Y. Guo and J. Zhang are with the School
of Mathematical Science, University of Science and Technology of China.
}
\thanks{$^\dagger$Corresponding author. Email: \texttt{juyong@ustc.edu.cn}.}}

\markboth{IEEE Transactions on Visualization and Computer Graphics}%
{Shell \MakeLowercase{\textit{et al.}}: Bare Demo of IEEEtran.cls for Computer Society Journals}

\IEEEtitleabstractindextext{%
\begin{abstract}
We introduce PICA, a novel representation for high-fidelity animatable clothed human avatars with physics-accurate dynamics, even for loose clothing. 
Previous neural rendering-based representations of animatable clothed humans typically employ a single model to represent both the clothing and the underlying body.
While efficient, these approaches often fail to accurately represent complex garment dynamics, leading to incorrect deformations and noticeable rendering artifacts, especially for sliding or loose garments.
Furthermore, previous works represent garment dynamics as pose-dependent deformations and facilitate novel pose animations in a data-driven manner. This often results in outcomes that do not faithfully represent the mechanics of motion and are prone to generating artifacts in out-of-distribution poses.
To address these issues, we adopt two individual 3D Gaussian Splatting (3DGS) models  with different deformation characteristics, modeling the human body and clothing separately. This distinction allows for better handling of their respective motion characteristics. 
With this representation, we integrate a graph neural network (GNN)-based clothed body physics simulation module to ensure an accurate representation of clothing dynamics.
Our method, through its carefully designed features, achieves high-fidelity rendering of clothed human bodies in complex and novel driving poses, significantly outperforming previous methods under the same settings.
\end{abstract}

}
\maketitle

\IEEEdisplaynontitleabstractindextext
\begin{figure*}[h]
    \centering
    \includegraphics[width=\textwidth]{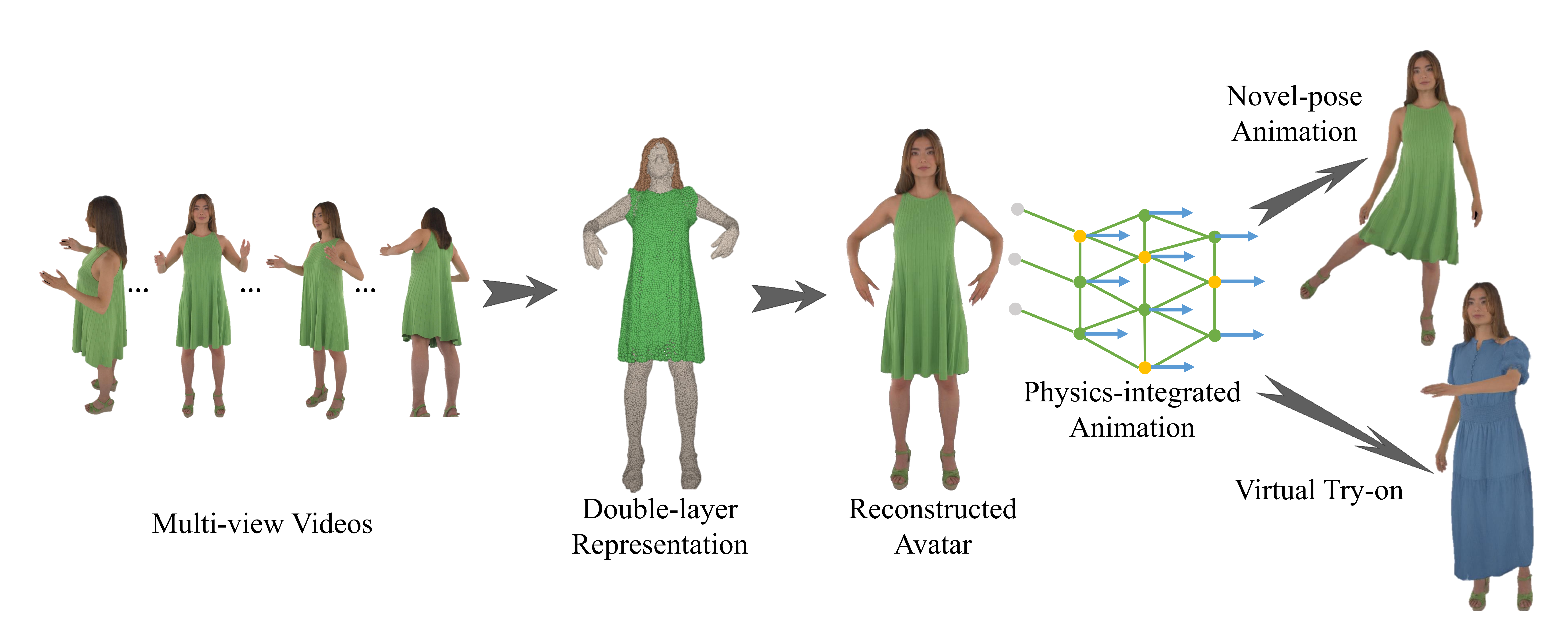}
    \vspace{-10mm}
    \caption{Given multi-view RGB inputs (25 views in this example), PICA reconstructs the clothed avatar as a double-layer representation and applies novel pose animation and virtual try-on with a physics-integrated driving module.}
    \label{fig:teaser}
    \vspace{-3mm}
\end{figure*}
\IEEEpeerreviewmaketitle

\IEEEraisesectionheading{\section{Introduction}\label{sec:introduction}}
\IEEEPARstart{H}{igh}-fidelity animation and rendering of clothed human performers is an important research problem in computer graphics and 3D computer vision, with broad applications in VR/AR, holographic communication, and other areas.
Although this topic has been studied for a long time, it remains challenging due to the complex non-rigid deformation of the clothing, especially in novel poses. The key challenge is how to accurately model loose clothing and effectively generate realistic and physically faithful clothing dynamics.

 The majority of human avatar reconstruction and animation methods have employed parametric body models such as SMPL~\cite{SMPL:2015} to reconstruct explicit 3D meshes from monocular~\cite{jiang2022selfrecon,chen2024neural,alldieck2018video} or multi-view~\cite{peng2024animatable} RGB videos, and integrate neural textures~\cite{liu2021neural,habermann2021real} or neural implicit fields~\cite{peng2021neural, weng_humannerf_2022_cvpr, peng2021animatable,li2022tava,habermann2023hdhumans, wang2022arah} for enhanced realistic rendering. 
Recently, 3DGS~\cite{kerbl20233d} proposes to model static scenes with 3D Gaussians. This method not only facilitates rapid training and inference but also delivers superior rendering quality compared to neural radiance field representation (NeRF)~\cite{mildenhall2021nerf}.
Subsequent studies~\cite{jena2023splatarmor,lei2023gart,moreau2023human} have explored integrating 3DGS with human priors~\cite{SMPL:2015,SMPL-X:2019} to reconstruct high-fidelity human avatars.
While effective in rendering human avatars in training poses, they struggle to explicitly control the motions of the avatar and generate lifelike animations.
Since most of these methods treat clothed avatars as a single layer based on SMPL (or SMPL-X) meshes, it is difficult for these methods to accurately represent and render the complex deformations of loose-fitting clothing.
Moreover, most methods animate the avatar only through a skeleton of parametric models using Linear Blend Skinning (LBS), leading to unconvincing results in novel poses.
Some methods~\cite{li2024animatablegaussians,Zielonka2023Drivable3D,lin2024layga} attempt to model the non-rigid deformation of clothing as pose-dependent deformation, and apply novel pose driving in a data-driven manner. A key problem with these methods is that clothing can be in different states even when the human body is in the same pose, so modeling the dynamics of the clothing as a function of the single-frame human pose may produce motion results that are not faithful to the mechanics of the motion.
In addition, this can lead to unreasonable results in out-of-distribution poses, requiring long training sequences that cover most poses.

To address these challenges, we introduce PICA, a novel representation for high-fidelity, animatable clothed human avatars with physically accurate dynamics. 
PICA is capable of synthesizing high-fidelity novel views of human performance in novel driving poses. 
This is achieved by modeling the clothes and the underlying body separately, and incorporating a physics-based driving module that accurately captures the dynamics of the garment. 
To effectively represent the diverse deformation characteristics of clothing and the human body, we combine explicit mesh and dynamic 3DGS to reconstruct the human body and clothing respectively. Based on this double-layer representation, we then utilize the graph neural network to predict the acceleration of each frame of clothes to ensure dynamic and realistic clothing movement effects.

Specifically, PICA employs two sets of dynamic 3D Gaussians, each attached to corresponding explicit meshes, to model the dynamic nature of the body and its clothing, respectively.  
We also use the mesh-aligned Gaussian representation like~\cite{waczynska2024games} to align the Gaussians with the underlying meshes.
To accurately distinguish the clothing from the naked body, we introduce a clothing segmentation term alongside a geometry regularization term. These terms help refine the shape of the explicit meshes, ensuring that both the clothing mesh and the corresponding 3D Gaussians remain completely outside the human body.

After reconstructing the dynamic Gaussians-represented avatar in canonical space, we then animate the clothed avatar by driving the explicit meshes with a graph neural network (GNN)~\cite{grigorev2023hood} trained with a self-supervised physical loss. 
Compared to other neural simulation methods, HOOD~\cite{grigorev2023hood} generates a hierarchical graph from the reconstructed clothing and body meshes, and predicts the acceleration of each clothing node in the graph at successive time steps. 
This allows generalization to new garment types and shapes, which are not seen during training.
To ensure that the physics-driven results better match with the input video, we fine-tune the physical properties of the clothing mesh, such as mass and bending coefficient. 
Once the novel pose geometry sequence is obtained from HOOD, the 3D Gaussians are updated synchronously along with the explicit clothing mesh. This manner is particularly effective at producing plausible novel pose controls, even when the training data contains a limited number of human poses.
Thanks to our double-layer representation of independently modeling the human body and clothing, PICA is also capable of generating realistic virtual try-on results, as shown in Fig.~\ref{fig:teaser}.

In summary, the technical contributions of this paper include the following aspects:
\begin{itemize}
\item We achieve hyper-realistic renderings by utilizing a double-layer 3DGS representation for the clothed body, which is instrumental in facilitating separate animation of the clothing and body.
\item By integrating the double-layer representation with an efficient neural simulation model, PICA achieves satisfying animation results of clothed humans in novel views and novel poses.
\item Leveraging our dual-layer representation and physics-based driving module, PICA supports realistic virtual try-on and can animate avatars wearing novel clothing, expanding the possibilities for dynamic and realistic fashion presentations.
\end{itemize}




\section{Related Work}
\subsection{3DGS-based Human Avatars}
3DGS (3D Gaussians splatting)~\cite{kerbl20233d} represents static scenes as sets of 3D Gaussians. It could achieve fast training and real-time rendering while maintaining high rendering quality, because 3D Gaussians can be rasterized efficiently compared to sampling rays used by neural radiance field (NeRF).
Recently, a large number of 3DGS-based works have demonstrated its potential.
GaMeS~\cite{waczynska2024games} proposes a mesh-aligned Gaussians representation where the reconstructed model is edited by directly editing the underlying mesh.
PhysGaussian~\cite{xie2023physgaussian} adds physical properties to 3D Gaussians and achieves high-quality novel motion synthesis through physically grounded Newtonian dynamics.

Recently, a line of works~\cite{jena2023splatarmor,lei2023gart,moreau2023human,kocabas2023hugs,zheng2023gpsgaussian,GauHuman,hu2024gaussianavatar,zhu2023ash,qian20233dgs,li2023human101} have modeled animatable human avatars using 3DGS.
However, most of these methods only represent both the human body and the clothing in a single layer, which is cumbersome for the reconstruction and animation of loose clothing.
To solve this problem, D3GA~\cite{Zielonka2023Drivable3D} models the human body and clothing separately, embeds 3D Gaussians in tetrahedral cages, and uses cage deformations to animate drivable avatars.
Animatable Gaussians~\cite{li2024animatablegaussians} proposes to learn a parametric template for loosely dressed performers, and uses a UNet to predict the properties of 3D Gaussians from posed position maps rendered from the template.
LayGA~\cite{lin2024layga} extends this representation to multi-layer for clothing transfer.
However, like most of these avatar methods, they model the non-rigid deformation of the clothes with a pose-conditioned MLP, which can lead to unrealistic cloth behavior in novel poses.
Our method, on the other hand, integrates physics-based driving to achieve reasonable and realistic animation results.

\subsection{Dynamic Clothes Modeling}
Modeling the motion of clothing along with the human body is a challenging problem.
Classical simulation methods~\cite{tang2018cloth,li2022diffcloth, Li2021CIPC} generate per-frame deformation with physics-based simulation, which can produce highly realistic results, but are computationally expensive.
To overcome this limitation, many researches\cite{bertiche2021deepsd,pan2022predicting,chentanez2020cloth,gundogdu2019garnet,jin2020pixel,pfaff2020learning,vidaurre2020fully} employ neural networks to predict clothing deformation from pose and shape parameters of the human body in a data-driven manner. 
Early works~\cite{chentanez2020cloth,pan2022predicting,vidaurre2020fully} typically require a long time to generate the training data by simulation and then train the network to predict garment deformation in a supervised manner.
Some works~\cite{santesteban2022snug,bertiche2020pbns} introduced a set of physics-based losses, transforming physics-based simulation into a self-supervised optimization problem.
Based on this, HOOD~\cite{grigorev2023hood} proposes to use hierarchical graphs and graph neural networks to enable real-time prediction for arbitrary garment types and body shapes.
In addition, all of the above works model the geometry simulation as a separate issue without modeling the realistic appearance.

Recently, some studies have been conducted to model a lifelike clothed human avatar by combining these simulation methods with neural rendering. 
Dressing Avatars~\cite{xiang2022dressing} introduces a neural clothing appearance model that operates on top of explicit geometry, which can obtain human and clothing appearance through high-fidelity tracking during training and apply novel pose driving by physically simulated geometry.
Concurrent with our work, Physavatar~\cite{PhysAavatar24} tracks the high fidelity mesh of each frame through 4D Gaussians and animates the avatar by driving the mesh with C-IPC~\cite{Li2021CIPC}. 
While achieving high-fidelity results for novel pose synthesis, this approach requires a ground truth mesh for initialization and an extraordinary dense camera setup (160 views) for tracking. 
Meanwhile, their inference speed is slow due to the high computational cost of I-IPC~\cite{Li2021CIPC}. 
By comparison, our work requires only a rough template mesh and a relatively sparse input (25 views for all our examples) on the same dataset, and the inference speed can achieve around 12fps with the assistance of a pretrained clothing dynamics model HOOD~\cite{grigorev2023hood}.



\section{Method}
\begin{figure*}
    \centering
    \includegraphics[width=\textwidth]{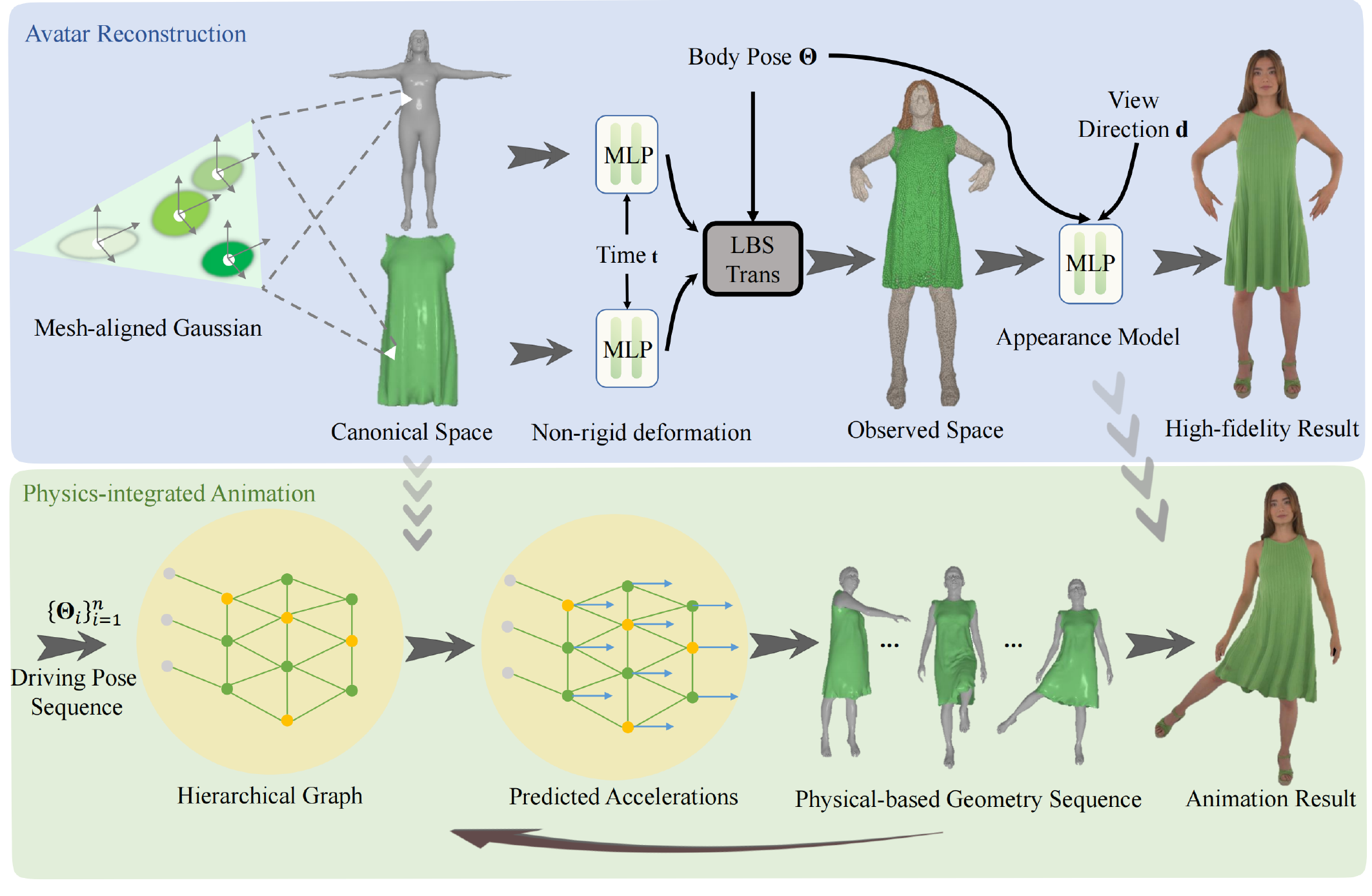}
    \vspace{-8mm}
    \caption{Overview. PICA represents clothed human avatars as two separate template meshes and corresponding mesh-aligned Gaussians. The avatar in canonical space is first deformed to observed space by non-rigid deformation and LBS, and then rasterized to the image space of the given camera with a pose-dependent color MLP. After reconstructing the avatar with appearance loss and geometry loss, PICA utilizes a hierarchical graph-based neural dynamics simulator to generate the simulation geometry sequence, which is rendered to the final animation result according to the trained appearance model.}
    \label{fig:pipeline}
    \vspace{-3mm}
\end{figure*}

Fig.~\ref{fig:pipeline} shows an overview of our method. 
Given multi-view videos of a human performer, our goal is to reconstruct the clothing and underlying body and then to animate the reconstructed clothed avatar with physically faithful clothing dynamics. 
To facilitate realistic animation of clothing driven by underlying body movements, we introduce a double-layer 3DGS representation that models the body and clothing separately. 
For efficient and physically accurate clothing simulation, we utilize a hierarchical graph-based neural dynamics simulator tailored for various garments. 
With the double-layer representation and the neural physics simulator, our method achieves high-fidelity novel view and novel pose synthesis results with natural clothing dynamics. 

We first provide some background in Sec.~\ref{sec:background}, and then introduce the construction of the clothed human avatar in Sec.~\ref{sec:representation}, in which the training losses are illustrated in Sec.~\ref{sec:loss}. Finally, we introduce the physics-based driving module in Sec.~\ref{sec:physics_deform} and the implementation details in Sec.~\ref{sec:detail}.

\subsection{Background}
\label{sec:background}
3DGS~\cite{kerbl20233d} represents a static 3D scene as a set of 3D Gaussians. Each 3D Gaussian is defined by its mean position $\mathbf{x}$, rotation matrix $\mathbf{R}$, scaling matrix $\mathbf{S}$, opacity $\alpha$ and SH coefficients $\mathbf{f}$. The 3D covariance matrix $\Sigma$ is then computed by:
\begin{equation}
\Sigma=\mathbf{R} \mathbf{S} \mathbf{S}^T \mathbf{R}^T.
\end{equation}
The images are then rendered based on the 3D Gaussians using a differential rasterization and splatting strategy.

In practice, the 3D Gaussians are first projected onto the 2D image plane. Given a view transformation matrix $\mathbf{W}$, the covariance matrix $\Sigma^{\prime}$ in 2D coordinates is given by:
\begin{equation}
\Sigma^{\prime}=J W \Sigma W^T J^T,
\end{equation}
where $J$ is the Jacobian of the affine approximation of the projective transformations.
After projecting the 3D Gaussians onto the image plane, a point-based $\alpha$-blending is applied to compute the color of each pixel, sorted by the depth of each Gaussian:
\begin{equation}
\label{eq:render}
C=\sum_i\left(\alpha_i^{\prime} \prod_{j=1}^{i-1}\left(1-\alpha_j^{\prime}\right)\right) c_i,
\end{equation}
where $\alpha_i^{\prime}$ is the weighted opacity according to the probability density of the i-th projected 2D Gaussian at the target pixel. 
$c_i$ is the view-dependent color computed from the view direction $d$ and the SH coefficients $f_i$. The 3D Gaussians are then optimized by a photometric loss. During training, 3DGS utilizes a self-adaptive densification and pruning strategy to control the number of 3D Gaussians.

\subsection{Avatar Representation}
\label{sec:representation}
\subsubsection{Avatar in canonical space}
\label{sec:canonical representation}
We begin by representing our avatar in canonical space. To drive the 3D Gaussians, we represent our avatar as 3D Gaussians anchored to the faces of the template meshes $\{M^{\textrm{body}},M^{\textrm{clothing}}\}$, where $M^{\textrm{body}}$ and $M^{\textrm{clothing}}$ are the body and clothing meshes, respectively. 
In practice, the body mesh is initialized directly as the template mesh of the SMPL-X model~\cite{SMPL-X:2019}. 
For people with long hair, we anchored some additional triangular faces behind the head to model the hair.
To get the clothing mesh in canonical space, we first need to get a full body mesh. 
For tight clothing, we also use the SMPL-X mesh directly. 
For loose clothing or dresses, we reconstruct a template mesh from a single frame using the method provided by Animatable Gaussians~\cite{li2024animatablegaussians}.
The 2D segmentation masks are projected onto the full body mesh to obtain $M^{\textrm{clothing}}$.

We then sample 3D Gaussians on the faces of $M^{\textrm{body}}$ and $M^{\textrm{clothing}}$. 
Since we aim to animate the avatars by driving the template mesh with the physical prior, we define the geometry properties of the Gaussians (mean position, scaling, and rotation) solely in terms of the mesh properties.
Specifically, given the position of the vertices ${p}_{i j}$ of the i-th face (j=1,2,3), we can define the mean $\mathbf{x}$ of the 3D Gaussians anchored to that face in terms of their barycentric coordinates $b_j$ and normal offset $\delta$:
\begin{equation}
\mathbf{x}=\sum_{j=1}^3 b_j \mathbf{p}_{i j} + \delta \mathbf{n_i},
\label{formula: bary}
\end{equation}
where $\mathbf{n_i}$ is the normal vector of the face.

To make the rendered surface consistent with the surface of the template meshes, we want the 3D Gaussian to be as flat as possible. 
Therefore, we use the representation similar to GaMeS~\cite{waczynska2024games} to align the Gaussian and triangular faces of the template meshes. 
Specifically, we first compute the orthonormal bases [$\mathbf{r_1}$,$\mathbf{r_2}$,$\mathbf{r_3}$] of the i-th face, where $\mathbf{r_1}$ = $\mathbf{n_i}$ is given by the face normal. Then we define
\begin{equation}
\begin{gathered}
\mathbf{r_2}=\frac{\left(\mathbf{p}_{i1}-\mathbf{p}_{i0}\right)}
{\left\|\mathbf{p}_{i1}-\mathbf{p}_{i0}\right\|}, \quad \mathbf{r}_{3} = \mathbf{r}_{1} \times \mathbf{r}_{2},
\end{gathered}
\end{equation}
where $\mathbf{p}_{i0}$ is the center of the i-th triangular face. The rotation matrix of the i-th face and the Gaussians anchored on it can be defined as $R$ = [$\mathbf{r_1}$,$\mathbf{r_2}$,$\mathbf{r_3}$], which makes sure the Gaussians are aligned with the face.

To make the Gaussian kernels as flat as possible, the scales of the Gaussians are calculated by:
\begin{equation}
S = [\epsilon, \left\|\mathbf{p}_{i1}-\mathbf{p}_{i0}\right\| \cdot s_2, \left\|\mathbf{p}_{i2}-\mathbf{p}_{i0}\right\| \cdot s_3],
\end{equation}
where $\epsilon$ is a small constant number. Since the first scaling parameter corresponds to the face normal, $\{s_i\}_{i=2}^3$ is the per-Gaussian optimizable parameter.


\subsubsection{Deformation}
We then translate the 3D Gaussians from canonical space to observed space for reconstruction by driving $M^{\textrm{body}}$ and $M^{\textrm{clothing}}$. The deformation consists of non-rigid deformation by tiny MLPs and rigid deformation by LBS. 
Formally, for a vertex $\mathbf{p}$ of the template meshes in canonical space, we obtain the transformed vertex $\mathbf{p}^\prime$ by:
\begin{equation}
\label{dfm}
\begin{gathered}
\mathbf{p}^\prime= \sum_{i=1}^N w_i G_i(\boldsymbol{\theta}, \boldsymbol{\beta}) \cdot (\mathbf{p}+\Delta \mathbf{p}),\quad 
\Delta \mathbf{p}= f_{\textrm{part}} (\mathbf{p},t),
\end{gathered}
\end{equation}
where $w_i$ is the blend weight of $v$, $G_i(\boldsymbol{\theta}, \boldsymbol{\beta})$ is the transformation matrix of the i-th bone given the SMPL-X parameters $\boldsymbol{\theta}$ and $\boldsymbol{\beta}$.
The blend weights of the clothing mesh are initialized with the nearest neighbor of the body mesh, and are jointly optimized during training.
$\Delta p$ is the non-rigid deformation predicted by an MLP $f_{\textrm{part}}$, which is conditioned on the frame index $t$.
We use two MLPs to better model the non-rigid deformation of the body and clothing separately. 
The geometry properties of the 3D Gaussians are then updated using the formula defined in Sec.~\ref{sec:canonical representation}.

\subsubsection{Appearance}
The original 3DGS uses spherical harmonics to represent color, which works well for static scenes. However, it needs pose-dependent color to model self-shadows and wrinkles in garments. Therefore, we use a color MLP to pre-compute the color from the per-Gaussian feature vector, view direction, and body pose parameters, where the color of each 3D Gaussian is computed by:
\begin{equation}
\label{formula: color}
\mathbf{c_i} = f_{\textrm{color}} \left(f_i, \boldsymbol{\theta}, h_t, d^\prime \right),
\end{equation}
where $f_i$ is the feature vector of each Gaussian, and $\boldsymbol{\theta}$ refers to the body pose parameters of the SMPL-X model.
We use the pose projection strategy proposed in Animatable Gaussians~\cite{li2024animatablegaussians} for better generalization to novel poses.
$h_t$ denotes an embedding vector with the frame index $t$~\cite{peng2021neural}, which is set to encode the time-varying factors.
Similar to~\cite{xie2023physgaussian}, we canonicalize the view direction by:
\begin{equation}
\label{formula: rot}
d^\prime= R_i^T d,
\end{equation}
where $d$ is the view direction vector from the camera center to the i-th Gaussian and $R_i$ refers to the rotation matrix of the i-th Gaussian.

\subsection{Training}
\label{sec:loss}
We optimize $V^{\textrm{body}}$, $V^{\textrm{clothing}}$, the properties of the 3D Gaussians, the deformation MLPs, and the pose-dependent color MLP. Since the input SMPL-X parameters may be inaccurate, we also optimize the SMPL-X parameters as well as the blend weight of $V^{\textrm{clothing}}$. We train these parameters with the following loss function:
\begin{equation}
    L=L_{\text{color}} +  L_{\text{mask}}+  L_{\text{seg}} + L_{\text{opac}} +  L_{\text{geo}}.
\end{equation}
Here, the appearance loss $L_{\text{color}}$ is used to ensure rendering quality with $L_{2}$ term, SSIM term, and perceptual term~\cite{zhang2018unreasonable}:
\begin{equation}
L_{\text{color}}= L_{\text{mse}} + \lambda_{\text{ssim}} L_{\text{ssim}} + \lambda_{\text{lpips}} L_{\text{lpips}}.
\end{equation}
The mask loss $L_{\text{mask}}$ is the $L_{2}$ error between the rendered mask and the ground truth mask. To ensure that the clothing is represented by the Gaussians positioned on the clothing template mesh, we introduce a segmentation loss:
\begin{equation}
\begin{gathered}
L_{\text{seg}} = \lambda_{\text{seg}} BCE(\hat{\mathcal{L}},\mathcal{L}), \quad
    \hat{\mathcal{L}} = \sum_i\left(\alpha_i^{\prime} \prod_{j=1}^{i-1}\left(1-\alpha_j^{\prime}\right)\right) l_i,
\end{gathered}
\end{equation}
where $l_i=1$ if the Gaussian is anchored to $M^{\textrm{clothing}}$ otherwise 0, and $\mathcal{L}$ is the ground truth clothing segmentation generated by PerSAM~\cite{zhang2023personalize} and Grounded SAM\cite{ren2024grounded}.

In our representation, the Gaussian is flat and aligned with the mesh surface, and the human body and clothing are opaque in most cases. 
Therefore, we use an opacity sparse loss from \cite{lombardi2019neural} to encourage the opacity of the Gaussians to be either 0 or 1:
\begin{equation}
L_{\text{opac}}=\lambda_{\text{opac}} \frac{1}{N} \sum_{i=1}^{N}\left(\ln \left(o_{i}\right)+\ln \left(1-o_{i}\right)\right),
\end{equation}
where $o_{i}$ is the opacity of the i-th Gaussian. We also use a geometry loss to make sure the template meshes are smooth and regular, and that $M^{\textrm{clothing}}$ is outside of $M^{\textrm{body}}$:
\begin{equation}
    L_{\text{geo}} = L_{\text{laplacian}} + L_{\text{normal}} + L_{\text{collision}} + L_{\text{distance}},
\end{equation}
where $L_{\text{laplacian}}$ is the Laplacian loss and $L_{\text{normal}}$ is the normal consistency loss induced on the meshes, which ensure a smooth geometry and is inspired by Luan et al.~\cite{luan2021unified}. $L_{\textrm{collision}}$ refers to a collision penalty term from cloth simulation literature~\cite{sifakis2012fem,narain2012adaptive}, which is used to ensure that $M^{\textrm{clothing}}$ are in the outer normal direction of $M^{\textrm{body}}$:
\begin{equation}
L_{\text {collision }}=\lambda_{\text {collision }}  \frac{1}{n} \sum_{i=1}^{n}\max (\epsilon- (v_i - v_j) \cdot n_j, 0)^3,
\end{equation}
where $v_i$ is the i-th vertex of the clothing mesh, $v_j$ is the nearest neighbor of $v_i$ in the body mesh, and $n_j$ is the corresponding normal vector of $v_j$. $L_{\text{distance}}$ is used to prevent the body mesh from being far away from the SMPL-X initialization.

\subsection{Physics-based Driving}
\label{sec:physics_deform}
After reconstructing the double-layer avatar representation, we then drive the body through LBS and the clothing through physical simulation respectively. Following HOOD~\cite{grigorev2023hood}, we adopt a hierarchical graph generated from $M^{\textrm{body}}$ and $M^{\textrm{clothing}}$ to predict the per-frame acceleration of each vertex on $M^{\textrm{clothing}}$.
\begin{figure*}
    \centering
    \includegraphics[width=\textwidth]{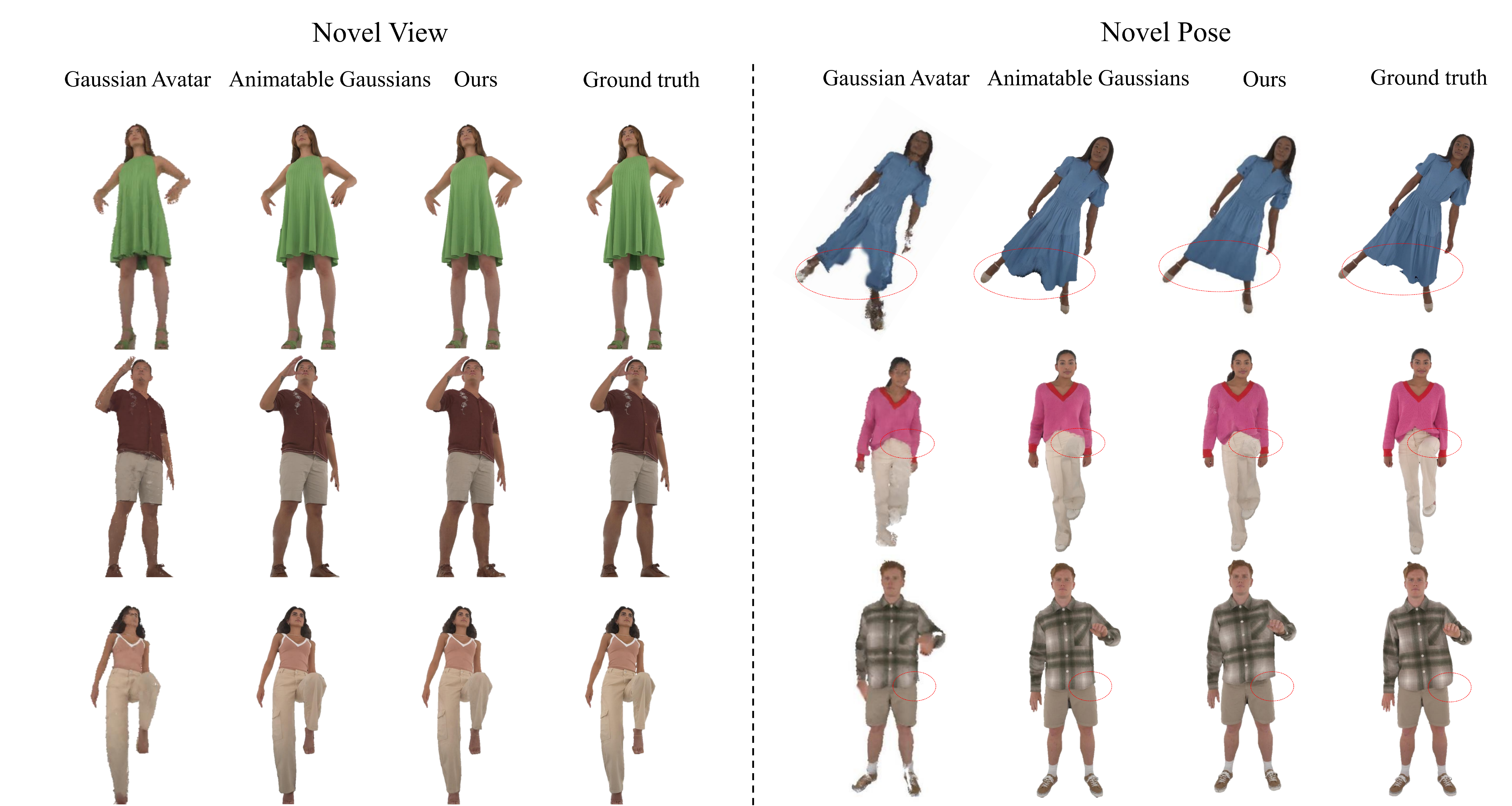}
    \vspace{-8mm}
    \caption{Qualitative results of novel view synthesis (left) on training frames and novel pose animation (right).}
    \label{fig:comp_view}
    \vspace{-5mm}
\end{figure*}

\subsubsection{Graph structure}
The hierarchical graph mainly consists of the vertices of $M^{\textrm{clothing}}$ and $M^{\textrm{body}}$ and the edges of $M^{\textrm{clothing}}$, extended by the body-clothing edges.
Formally, a hierarchical graph is defined as:
\begin{equation}
G=(V^{\textrm{body}}\cup V^{\textrm{clothing}}, E^{\textrm{clothing}}\cup E^{\textrm{coarse}}\cup E^{\textrm{clothing2body}}), 
\end{equation}
where $V^{\textrm{body}}$ and $V^{\textrm{clothing}}$ are the vertices of $M^{\textrm{body}}$ and $M^{\textrm{clothing}}$ respectively, $E^{\textrm{clothing}}$ are the edges of $M^{\textrm{clothing}}$. 
$E^{\textrm{coarse}}$ are the edges of the simplified mesh of $M^{\textrm{clothing}}$, designed for long-range connections between vertices. $E^{\textrm{clothing2body}}$ are the edges from $M^{\textrm{clothing}}$ to its nearest vertex on $M^{\textrm{body}}$ if the distance is below a threshold. Note that $E^{\textrm{clothing2body}}$ is calculated every frame based on the current positions of $V^{\textrm{clothing}}$ and $V^{\textrm{body}}$.

The vertices and edges of $G$ are anchored by feature vectors. 
Nodal features consist of current state variables (position $p_t$, velocity $v_t$, vertex normal 
$n_t$) and physical properties $\rho$ (mass, bending coefficient, etc.). Edge features store the relative position between its two adjacent nodes. 
Please refer to HOOD~\cite{grigorev2023hood} for more details.
For simplicity, we denote the state (the vertices, the edges and the corresponding features) of the hierarchical graph at time $t$ as $G_{t}$.

\subsubsection{Acceleration Regression}
To better match with the training data, we initialize $G_{0}$ using Eq.~\ref{dfm}. 
Then we update the position of $M^{\textrm{clothing}}$ and the state of $G$ by:
\begin{equation}
    \label{eq:gnn}
    \begin{gathered}
        a_t = GNN(G_{t-1},\rho),\quad
        v_t = v_{t-1} + a_t,\\
        p_t = p_t + v_t,\quad
        G_{t} = \Omega(G_{t-1}, p_t,v_t),
    \end{gathered}
\end{equation}

where $a$, $v$ and $p$ are the acceleration, velocity and position of $V^{\textrm{clothing}}$, respectively. 
$GNN(\cdot)$ refers to a graph neural network based on MeshGraphNets~\cite{pfaff2020learning}, which we use the pre-trained model from HOOD.
$\Omega(\cdot)$ means to update $G$ and the corresponding feature vectors according to the information of the current frame. 
$\rho \in \mathbb{R}^4$ refers to the physical properties of the garment, which is an optimizable parameter according to the reconstructed results:
\begin{equation}
\rho =\mathop{\arg\min}_{\rho} \ \ \sum_{t=0}^{T}\left\|p_{t+1}^{r}-p_{t+1}^{s}(p_{t}^{r}, p_{t-1}^{r}, \rho)\right\|_{2}^{2}.
\end{equation}
Where $p_{t}^{r}$ is the reconstructed position of the vertices of the clothing mesh at frame $t$, which is calculated by Eq.~\ref{dfm}. 
$p_{t+1}^{s}$ refers to the simulated position by Eq.~\ref{eq:gnn} when given the physical properties $\rho$ and the graph state calculated from the reconstructed position $p_{t}^{r}$ and $p_{t-1}^{r}$.
We can now transfer the 3D Gaussians from the canonical space to the current space using the formulas defined in Sec.~\ref{sec:canonical representation}, and generate the novel pose driving and novel view synthesis results through Eq.~\ref{eq:render}.


\subsection{Implementation Details}
\label{sec:detail}
In practice, we anchor 13 Gaussians on each triangular face.
We trained our model on multi-view videos captured with 25 evenly distributed cameras from the ActorsHQ~\cite{isik2023humanrf} dataset.
For 1494$\times$1022 multi-view videos with 200 frames, we need about 40K iterations to converge, and it takes about 100 minutes with a single NVIDIA A100 GPU. During training, an annealing strategy is adopted for the pose input and the latent code to avoid the overfitting of the color net.
For evaluation, we animate the human body with LBS and simulate the garment dynamics with HOOD. The whole system runs at 12fps with a single NVIDIA A100 GPU.

\section{Experiments}
To demonstrate the effectiveness of our method, we compare with the state-of-the-art methods~\cite{li2024animatablegaussians,hu2024gaussianavatar} for novel view synthesis and novel pose animation.
We first perform a quantitative comparison for novel view synthesis to evaluate our avatar representation and reconstruction algorithm, and then qualitatively compare the results of novel pose animation.
Some ablation studies are also discussed to evaluate the necessity of our modules.
Finally, we show that our method supports realistic virtual try-on and can animate avatars wearing novel clothing.

\vspace{-1mm}
\begin{table}[h]
\caption{Quantitative comparison for novel view synthesis.}
\vspace{-3mm}
\label{table:compare}
\resizebox{\linewidth}{!}{
\begin{tabular}{cccccc}
\hline
Subject                  & Metric                  & Animatable Gaussians & GaussianAvatar & Ours             \\ \hline
\multirow{3}*{Actor01}    & PSNR/dB$\uparrow$            & \textbf{31.2204}     &      27.6649       & 31.1292          \\
                         & SSIM$\uparrow$            & 0.9506               &     0.9321    & \textbf{0.9588}  \\
                          & LPIPS$\downarrow$         & \textbf{0.0567}      &     0.1080   & 0.0642           \\ \hline
\multirow{3}{*}{Actor02} & PSNR/dB$\uparrow$            & \textbf{31.7390}     &    26.6044  & 30.3025          \\
                         & SSIM$\uparrow$             & 0.9598               &    0.9358     & \textbf{0.9636}  \\
                         & LPIPS$\downarrow$         & \textbf{0.0602}      &    0.1114     & 0.0716           \\ \hline
\multirow{3}{*}{Actor03} & PSNR/dB$\uparrow$           & 30.2085     &             24.7836    & \textbf{31.1996}          \\
                         & SSIM$\uparrow$             & 0.9246               &    0.8983    & \textbf{0.9341}  \\
                         & LPIPS$\downarrow$         & \textbf{0.0887}      &    0.1399    & 0.0788           \\ \hline
\multirow{3}{*}{Actor04} & PSNR/dB$\uparrow$           & 30.3622              &   27.2970    & \textbf{31.0899} \\
                         & SSIM$\uparrow$            & 0.9423               &   0.9280    & \textbf{0.9462}  \\
                         & LPIPS$\downarrow$          & \textbf{0.0530}      &  0.1083       & 0.0679           \\ \hline
\multirow{3}{*}{Actor06} & PSNR/dB$\uparrow$            & 30.1377              & 26.9535  & \textbf{30.3984} \\
                         & SSIM$\uparrow$             & 0.9607               &  0.9316       & \textbf{0.9668}  \\
                         & LPIPS$\downarrow$         & \textbf{0.0583}      &   0.1088     & 0.0679           \\ \hline
\multirow{3}{*}{Actor08} & PSNR/dB$\uparrow$           & \textbf{29.7758}     &    19.3875     & 29.1310          \\
                         & SSIM$\uparrow$            & 0.9429               & 0.8512      & \textbf{0.9469}  \\
                         & LPIPS$\downarrow$         & \textbf{0.0714}      &   0.1890     & 0.0860           \\ \hline
\multirow{3}{*}{Mean}    & PSNR/dB$\uparrow$             & \textbf{30.5739}     &    25.4485   & 30.5418          \\
                         & SSIM$\uparrow$            & 0.9468               &  0.9376   & \textbf{0.9527}  \\
                         & LPIPS$\downarrow$        & \textbf{0.0647}      &    0.1094    & 0.0727           \\ \hline
                         & \text{Training time}$\downarrow$  & $\sim$10h      &     $\sim$10h            &  $\sim$\textbf{2h}              \\
                         & \text{Inference speed}$\uparrow$ & $\sim$10fps    &   $\sim$7fps   &  $\sim$\textbf{20fps}                 \\ \hline
\end{tabular}
}
\end{table}
\vspace{-12mm}

\subsection{Dataset}
\label{data}

We use the ActorsHQ~\cite{isik2023humanrf} dataset for comparison, and perform experiments on six actors (A01, A02, A03, A04, A06, A08) with relatively loosing clothing. 
For each subject, we use the RGB images and corresponding foreground masks from the original dataset and the SMPL-X parameters provided by Animatable Gaussians~\cite{li2024animatablegaussians}, and generate the clothing segmentation with the help of PerSAM\cite{zhang2023personalize} and Grounded SAM\cite{ren2024grounded}.
\subsection{Comparison}

\subsubsection{Baselines.}
We compare with state-of-the-art methods based on 3DGS for novel view synthesis. Among them, Animatable Gaussians~\cite{li2024animatablegaussians} employs a powerful StyleGAN-based CNN to learn the pose-dependent Gaussian maps from posed position maps for modeling detailed dynamic appearances.
Gaussian Avatar~\cite{hu2024gaussianavatar} relies on a pose-dependent neural network to predict the parameters of 3D Gaussians.
We use the official open source code of these methods and extend Gaussian Avatar to our dense-view setting for equal comparison with our method. We select 25 evenly distributed views for training and other views for evaluation. For each actor, we choose 200 frames with relatively large motion for training.

\begin{figure*}
    \centering
    \includegraphics[width=\textwidth]{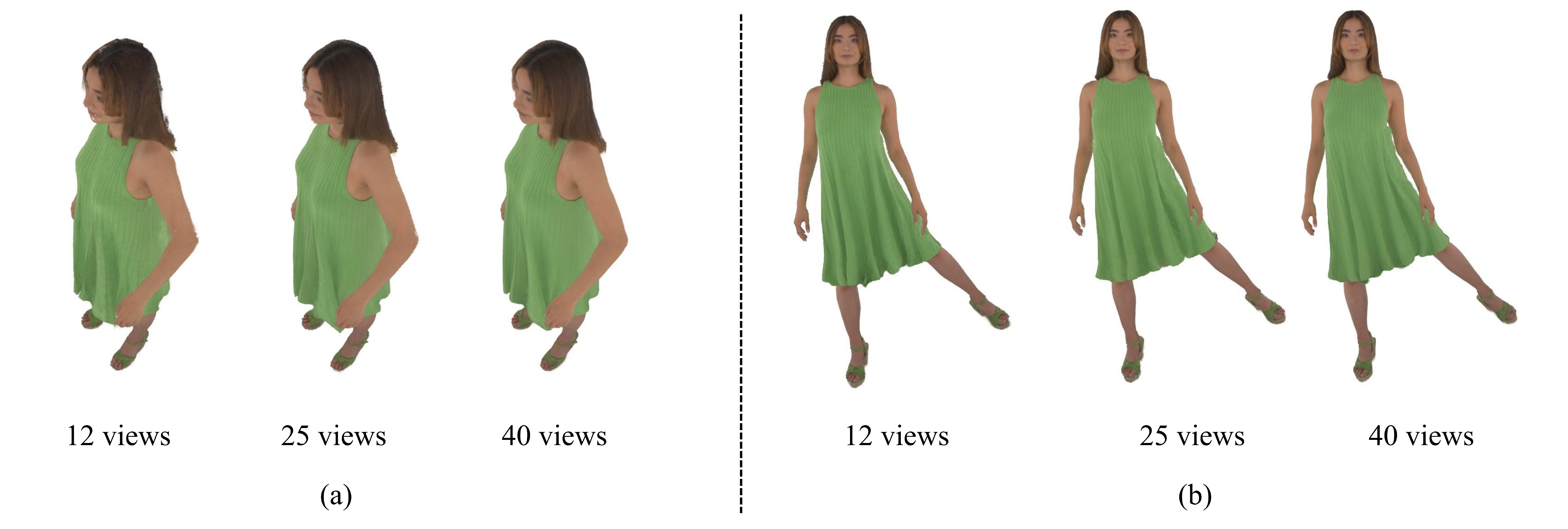}
    \vspace{-8mm}
    \caption{Ablation study on the number of training views. (a) Novel view synthesis results on the training frame. (b) Novel pose animation results on the training view.}
    \label{fig:cam_num}
    \vspace{-3mm}
\end{figure*}

\subsubsection{Novel View Synthesis.}
Following Animatable Gaussians, we use three standard metrics to quantify the results of novel view synthesis: peak signal-to-noise ratio (PSNR), structural similarity index (SSIM), and the learned perceptual image patch similarity (LPIPS)~\cite{zhang2018unreasonable}. 

\begin{figure*}
    \centering
    \includegraphics[width=\linewidth]{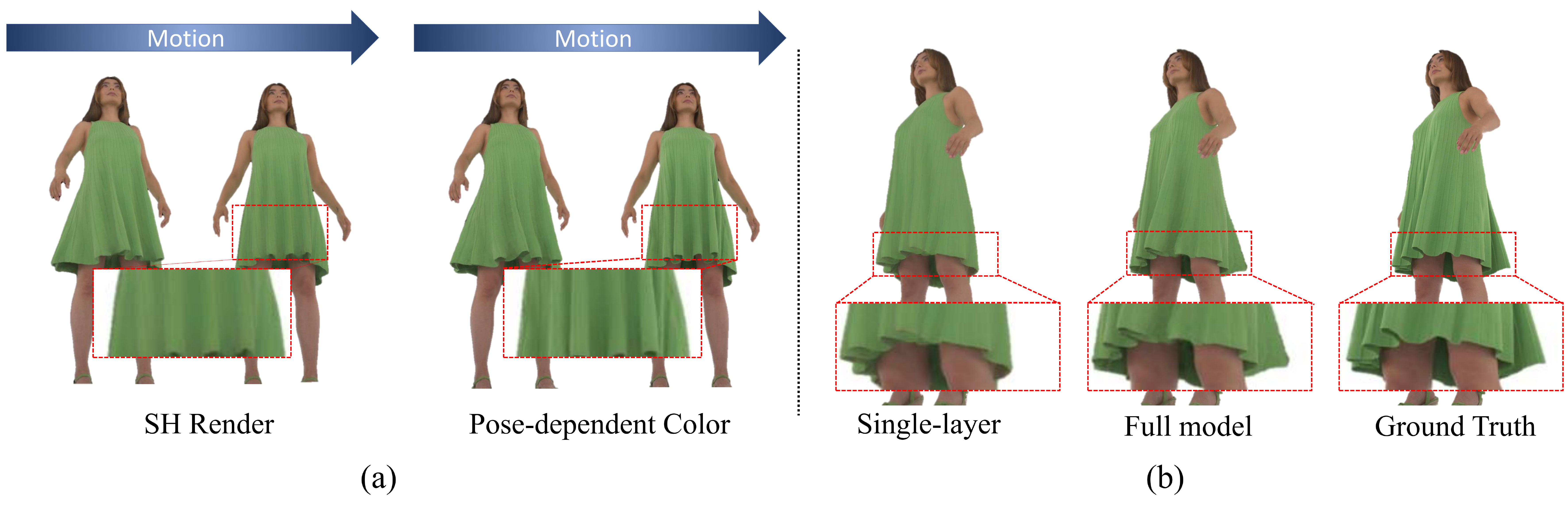}
    \vspace{-8mm}
    \caption{(a) Ablation study on the pose-dependent color. Compared to SH, our appearance model can better model self-shadows and wrinkles in garments. (b) Ablation study on the two-layer representation. The single-layer representation struggles to model the loose or sliding dynamics of clothing.}
    \label{fig:ablation_pose_and_layer}
\end{figure*}

The quantitative results are presented in Tab.~\ref{table:compare}. 
Our method outperforms Gaussian Avatar for all subjects and all metrics. 
Animatable Gaussians outperform us by a small margin for LPIPS.
Note that they rely on a powerful but computationally expensive StyleGAN-based CNN, resulting in slow training and inference speed compared to out method.
Fig.~\ref{fig:comp_view} demonstrates that both our method and Animatable Gaussians can generate high-fidelity novel view synthesis results, while the results from Gaussian Avatar appear to be blurry.

\subsubsection{Novel Pose Animation.}
For the novel pose animation, we only performed qualitative experiments since the clothing dynamic is not fully determined by human movements.
For evaluation, we directly use the pose sequence from the ActorsHQ dataset, which is not seen during training.
As shown in Fig.~\ref{fig:comp_view}, with the help of the physics-based driving module, PICA is able to generate novel pose animation results with physically faithful clothing dynamics, while the baseline methods fail to do so when the driving pose is far from the training set.

\vspace{-1mm}
\begin{table}[h]
\caption{Quantitative results of ablation study.}
\label{table2}
\vspace{-3mm}
\begin{tabular}{|lccc|}
\hline
Metric:                                      & PSNR/dB$\uparrow$                & SSIM$\downarrow$                  & LPIPS$\downarrow$                      \\ \hline
Single-layer representation &        29.74              &       0.949               &       0.796                \\ \hline
w/o pose-dependent color    &        29.32          &            0.951          &    0.753                   \\ \hline
w/o $L_{\text{seg}}$       &         30.14             &         0.953             &     0.750                  \\ \hline
w/o $L_{\text{geo}}$       & 24.89  & 0.924& 0.963\\ \hline
Full Model (12views)        & 27.24 & 0.931 & 0.912 \\ \hline
Full Model (25views)         & 30.13 & 0.953 & 0.751 \\ \hline
Full Model (40views)         & 31.25 & 0.959 & 0.698 \\ \hline
\end{tabular}
\end{table}
\vspace{-3mm}

\subsection{Ablation Study}
To further verify the effectiveness of the proposed modules in our method, we perform the following ablation studies. For evaluation, we use the data of one male and one female character (A1, A8), specifically selected for their loose clothing. 

\subsubsection{The number of training views}
We perform an ablation study to evaluate how the number of training views affects the quality of our novel view synthesis and novel pose animation. 
The results are shown in Tab.~\ref{table2} and Fig.~\ref{fig:cam_num}.
The number of views affects the appearance of the rendered avatars, since the data from the ActorsHQ dataset may not contain full-body information if the input is too sparse. It has little effect on the results of the novel pose animation, as PICA does not require much geometry detail to generate the physics-based geometry sequence with the prior from HOOD's pre-trained model. This demonstrates the potential of our method to support sparse even monocular input when the input data contains whole-body information.
\subsubsection{Double-layer representation}
We attempt to directly represent the clothes and the underlying body with a single layer mesh and corresponding Gaussians.
We use the SMPL-X mesh for A8 and the template mesh from Animatable Gaussians~\cite{peng2024animatable} for A1 as initialization.

\begin{figure*}
    \centering
    \includegraphics[width=\textwidth]{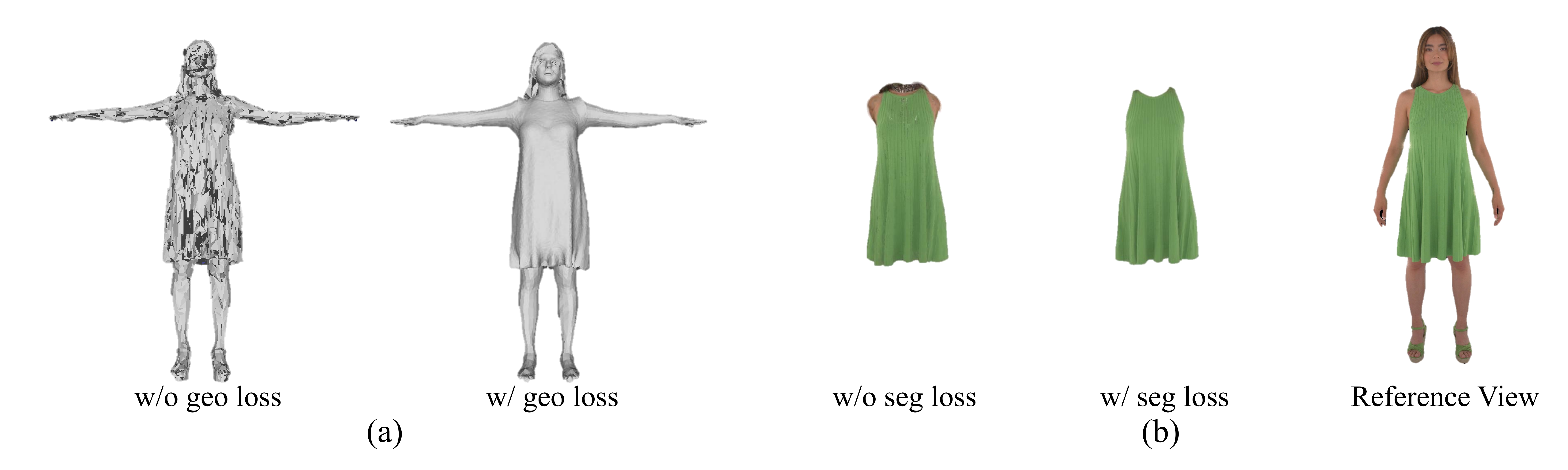}
    \vspace{-8mm}
    \caption{(a) Ablation study on geometry regularization loss. Without geometry loss, the topology of the template meshes becomes chaotic and the surfaces become noisy. (b) Ablation study on segmentation loss. With segmentation loss, PICA can better distinguish between body and clothing, resulting in more stable animation and virtual try-on results.}
    \label{fig:ablation_loss}
    \vspace{-3mm}
\end{figure*}

\begin{figure*}
    \centering
    \includegraphics[width=\textwidth]{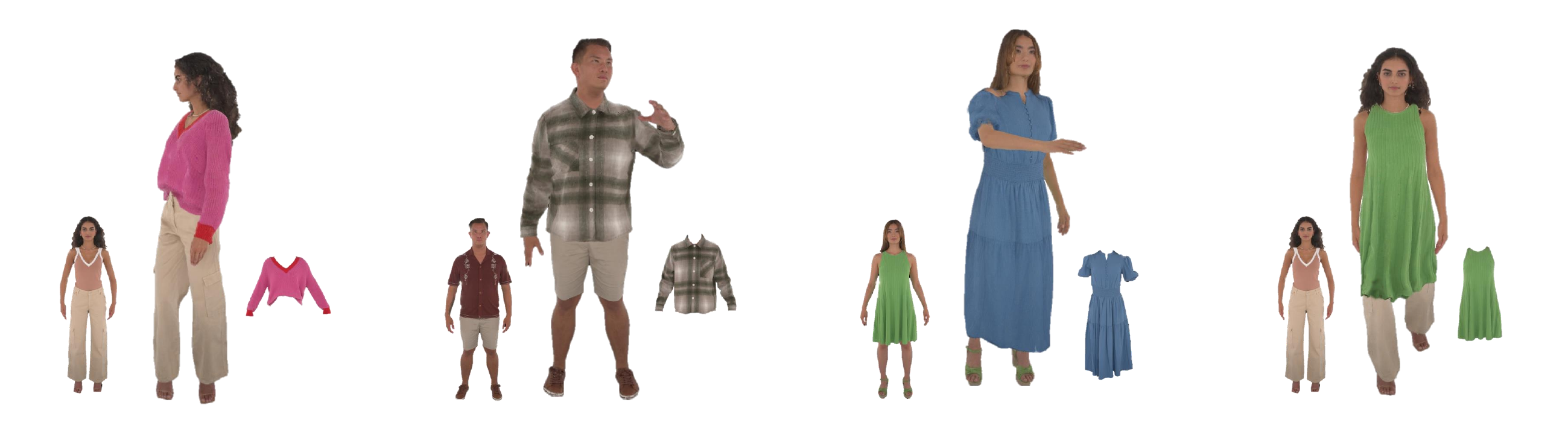}
    \vspace{-8mm}
    \caption{Results of Virtual try-on. PICA is capable of replacing the clothes and animating the newly dressed avatars. Please see to our video for dynamic results.}
    \label{fig:tryon}
    \vspace{-3mm}
\end{figure*}

As shown in Tab.~\ref{table2}, using two layers to model the human body and clothing separately provides further gains in the reconstruction of clothed avatars.
Fig.~\ref{fig:ablation_pose_and_layer} illustrates that the single-layer representation struggles to model proper garment sliding, and it fails to simultaneously model the motion of human body and loose clothes like long dress.
Moreover, the double-layer representation is important for subsequent clothing transfer and physics-based driving, that are difficult for the single-layer representation.

\subsubsection{Pose-dependent color}
To demonstrate the effectiveness of the pose-dependent color module, we attempt to use spherical harmonics to model the color of each Gaussian, as in the original 3D Gaussian.
As can be seen in Fig.~\ref{fig:ablation_pose_and_layer} and Tab.~\ref{table2}, it struggles to capture phenomena like self-shadows, which are pose-dependent, and turns to generate results with mean color.

\subsubsection{Segmentation Loss and Geometry Loss.}
We try to remove the segmentation loss and the geometry regularization loss from our training loss.
As shown in Fig.~\ref{fig:ablation_loss} and Tab.~\ref{table2}, although the segmentation loss cannot improve the rendering quality, it does help to distinguish the clothing Gaussians and the body Gaussians.
In addition, the geometry regularization loss ensures that the body mesh and clothing mesh are smooth and regular, which is necessary for subsequent physics-based driving.

\subsection{Application: Virtual Try-on}

Since PICA represent the human body and clothing separately, the trained model can be used for virtual try-on by simply replacing the clothing mesh and corresponding Gaussians.
We also apply a collision solver to ensure that the clothing mesh and Gaussians are outside the body, and then drive the human body with novel clothing in the same way as the standard pipeline. 
The results of the virtual try-on are shown in Fig.~\ref{fig:tryon} and in the accompanying video.

\section{Limitation and Future Work}
Although PICA can produce realistic novel pose animation results, fidelity cannot always be guaranteed even when optimizing physical parameters based on multi-view videos. 
Several factors lead to this limitation. First, most self-supervised neural cloth simulation methods, including HOOD, focus on making reasonable forward predictions rather than estimating inverse parameters based on real-world data.
Moreover, there is a gap between the rough template meshes in our work and the ground truth geometry, which may lead to errors in collision calculations.
Additionally, the state of the clothing is not solely determined by the body motion, so the problem itself is ambiguous.

In addition, since the internal body is completely occluded, we can only infer its geometric properties with body priors, while the appearance properties are completely unpredictable, which may lead to artifacts when the internal Gaussians are exposed during animations involving large motions. 
A potential solution is to utilize the general knowledge from a pre-trained 2D diffusion model as an appearance prior for the human body~\cite{kim2024gala}, which we leave as a future work.

\section{Conclusion}
We presented PICA, an efficient, physics-integrated novel pose animation method for clothed human performers, even for loose-fitting ones.
We introduced a double-layer representation for clothed avatars that allows for easy manipulation and editing capabilities while maintaining excellent visual fidelity.
Based on this novel and effective representation, PICA integrates a physics-based driving module to generate animation results with physically faithful clothing dynamics, and is capable of clothing transfer.
Extensive experimental results have demonstrated that PICA is capable of producing realistic and physics-faithful results for this challenging task.


\ifCLASSOPTIONcompsoc
  \section*{Acknowledgments}
\else
  \section*{Acknowledgment}
\fi
This work was supported by the National Natural Science Foundation of China (No. 62122071, No. 62272433) and the Youth Innovation Promotion Association CAS (No. 2018495).
\newpage

{
\bibliographystyle{IEEEtran}
\bibliography{PICA}
}

\end{document}